\documentclass[letterpaper, 10 pt, conference]{ieeeconf}  
\IEEEoverridecommandlockouts                              
\usepackage[cmex10]{amsmath}
\usepackage{graphics} 
\usepackage{epsfig} 
\usepackage{epstopdf}
\usepackage{mathptmx} 
\usepackage{times} 
\usepackage{amsmath} 
\usepackage{amssymb}  

\usepackage[table,xcdraw]{xcolor}
\usepackage{hyperref}
\usepackage{array}
\usepackage{bm}
\usepackage{cite}
\usepackage[font=small,labelfont=bf]{caption}
\usepackage{color}
\usepackage{setspace}
\usepackage[belowskip=-10pt]{caption}
\usepackage{mathptmx}
\DeclareMathAlphabet{\mathcal}{OMS}{cmsy}{m}{n}
\DeclareSymbolFont{largesymbols}{OMX}{cmex}{m}{n}
\usepackage[ruled,vlined]{algorithm2e}
\usepackage{multirow}
\title{\LARGE \bf
{DSL-Assembly: A Robust and Safe Assembly Strategy}
}

\author{Yi Liu$^{1, \dag}$ 
\thanks{$\dag$ denotes the corresponding author.}
\thanks{$^{1}$Y. Liu}
}

\begin{document}
\maketitle
\thispagestyle{empty}
\pagestyle{empty}

\begin{abstract}
A reinforcement learning (RL) based method that enables the robot to accomplish the assembly-type task with safety regulations is proposed. The overall strategy consists of grasping and assembly, and this paper mainly considers the assembly strategy. Force feedback is used instead of visual feedback to perceive the shape and direction of the hole in this paper. Furthermore, since the emergency stop is triggered when the force output is too large, a force-based dynamic safety lock (DSL) is proposed to limit the pressing force of the robot. Finally, we train and test the robot model with a simulator and build ablation experiments to illustrate the effectiveness of our method. The models are independently tested $500$ times in the simulator, and we get an $88.57\%$ success rate with a $4mm$ gap. These models are transferred to the real world and deployed on a real robot.
\textcolor{black}{We conducted independent tests and obtained a $79.63\%$ success rate with a $4mm$ gap.} 
Simulation environments: $\href{https://github.com/0707yiliu/peg-in-hole-with-RL}{\textbf{https://github.com/0707yiliu/peg-in-hole-with-RL}}$
\end{abstract}


\section{Introduction}
\label{sec:intro}

In some assembly-type tasks such as key insertion, humans complete the task through a brilliant combination of visual and contact force perception. However, it is challenging to endow robots with such capabilities, which require precise and rich graphical recognition algorithms and force perception-based algorithms. 

Visual feedback provides overall information about the object geometry and its surroundings used for pre-capturing and insertion. A purely vision-based model can be deployed on the robot to complete the partial assembly task \cite{intro-assemblyvision2}. But these models cannot make robots determine how much force is required.

Force sensing is a way for robots to determine physical parameters. Using it to provide partial feedback during collision or contact, the assembly process can be controlled accurately and safely \cite{intro-assemblyft1}. 
Therefore, the robot can obtain many environmental details by referring to multiple information such as vision and force sensing.

The application of RL \cite{rl-assembly2} provides an alternative policy for robots to complete the assembly-type tasks. The challenge with this policy is that the complete state cannot be observed instantly without observing the geometric model by the camera. Thus, the insertion policy not only tries to align the mismatch between targets but also needs to adjust the insertion direction \cite{ml-encode-vision1}. Further, the exploration of RL causes the robot to collide with the environment, hence, the safety of robot interaction deserves consideration.

\begin{figure}
    \centering
    \includegraphics[width=0.35\textwidth]{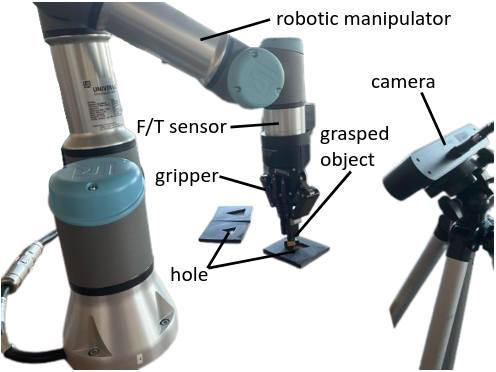}
    \caption{The overview of the robotic platform. }
    \label{fig:oversystem}
\end{figure}

In this paper, as shown in Fig. \ref{fig:oversystem}, we propose a strategy that can utilize multiple types of sensors with different 
characteristics. The proposed strategy is constructed by RL and can be generalized across similar manipulation tasks (e.g., similar geometries, configurations, and object sizes). 
The approach is to learn the joint representation of force/torque (F/T) sensors, robot proprioceptive information, and vision sensors through a fully connected neural network to obtain the necessary action and set the $\mathrm{DSL}$ for the robot. 

We summarize the key contributions as follows.
\begin{enumerate}
\item An RL containing multimodal information from which the insertion policy can be learned.
\item The DSL is set for the robot's motion trajectory to ensure safety interaction during the insertion process.
\item Simplifying the vision function by using F/T sensors to judge the precise position and direction of the hole instead of the camera.
\item Demonstrating effective use of the F/T sensor and visual feedback for hole search, alignment, and insertion. 
\end{enumerate}


\vspace{-5pt}
\section{Methodology}
\label{sec:method}

\subsection{Assembly Task Setting in Simulation}
\label{sec:sec:taskinsimulation}

\begin{figure}
    \centering
    \includegraphics[width=0.5\textwidth]{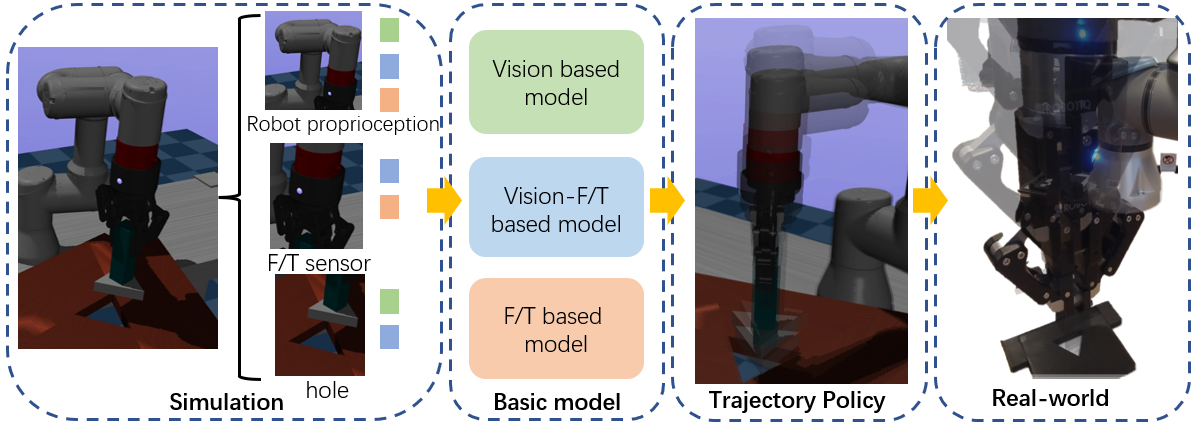}
    \caption{Model training schematic. The simulation part contains the state of the robot and the environment that we need, and the motion trajectories from the simulation can be transferred to the real robot.}
    \label{fig:system}
\end{figure}

As shown in Fig. \ref{fig:system}. Firstly, the training environment includes a 6-DoF robotic arm with a gripper, a workspace with a table, holes, and objects. Then, since this paper mainly focuses on the insertion action rather than the grasping action, the initial position of the robot end-effector (EEF) is fixed. The initial robot joint configuration is calculated via inverse kinematics. The target (hole) location is randomized within a defined domain that the EEF can reach. 

Finally, to encourage the robot to learn the policy effectively, we set the size of the object in the training environment to be constant and change the size of the hole, i.e., change the gap between the hole and the object.
Since the trajectory of the robot in this work is continuous, it is considered to use the standard PPO algorithm. Since we mainly consider the policy of insertion action as described above, the policy of grasping action is ignored in this policy and the object is fixed to the EEF in a way that it has 1-DoF.


\subsection{Observation, Action, and Reward}
\label{sec:sec:obs-act-rew}
The RL state $s_t$ consists of the robot state $s^r_t$, the assembly task state $s^a_t$. 
The robot state $s^r_t$ contains the robotic EEF position $\textbf{p}^{ee}_t=[x^{ee}_t, y^{ee}_t, z^{ee}_t]$, the EEF force and torque obtained with the F/T sensor $\textbf{f}^{ee}_t$, angular value of the last joint $q^6_t$ at the end of the robot. 
The task state $s^a_t$ contains the hole position $\textbf{p}^{h}_t=[x^{h}_t, y^{h}_t, z^{h}_t]$. 

The action $a_t$ consists of a three-dimensional displacement increment of the robot EEF $[\Delta x, \Delta y, \Delta z]$ and a rotation perpendicular to the insertion direction $\Delta \theta z$. 
For the $[\Delta x, \Delta y, \Delta z]$, the movement of the end is considered to compose the action set in order to satisfy the relative position $\textbf{r}^{ee}_h$ needed for the observation.

For the reward function $r_t$ required for the RL model, we choose the Euclidean distance as the basic function, i.e., the distance between the hole and the object. The observation $s_t$ does not include the position of the object, but fixes the object to the EEF (in \ref{sec:sec:taskinsimulation}), therefore, the Euclidean distance between the hole and the EEF $d^e_h$ is calculated. 
\begin{equation}
\begin{aligned}
r_t = &\sqrt{\alpha_1 * (x^{ee}_t - x^{h}_t)^2 
     + \alpha_2 * (y^{ee}_t - y^{h}_t)^2
       + \alpha_3 * (z^{ee}_t - z^{h}_t)^2} \\
    & + \alpha_4 * [d(\textbf{p}^{ee}_t, \textbf{p}^{h}_t) < \delta_1] \\
    & + \alpha_5 * z_{dist},
\end{aligned}
\label{eqn:rewardfunc}
\end{equation}
where $[\alpha_1, \alpha_2, \alpha_3]$ is the important weight vector of the different direction, $d(\cdot)$ expresses the Euclidean distance between $\textbf{p}^{ee}_t$ and $\textbf{p}^{h}_t$, this part could be activated when the distance less than the threshold $\delta_1$. $z_{dist}$ represents the distance in the insertion direction. The cost is defined as:
\begin{equation}
z_{dist}=\left\{
\begin{array}{rcl}
z^{h}_t - z^{ee}_t ,& & d(\textbf{p}^{ee}_t, \textbf{p}^{h}_t) < \delta_2\\
0 ,& & otherwise,
\end{array} \right.
\label{eqn:zdist}
\end{equation}
where $\delta_2$, similar to $\delta_1$,  is also the judgment threshold for distance $d(\cdot)$, but the value of $\delta_2$ is greater than the $\delta_1$ one, which means the reward can be activated when the object is inserted into the hole.


\begin{figure}
    \centering
    \includegraphics[width=0.38\textwidth]{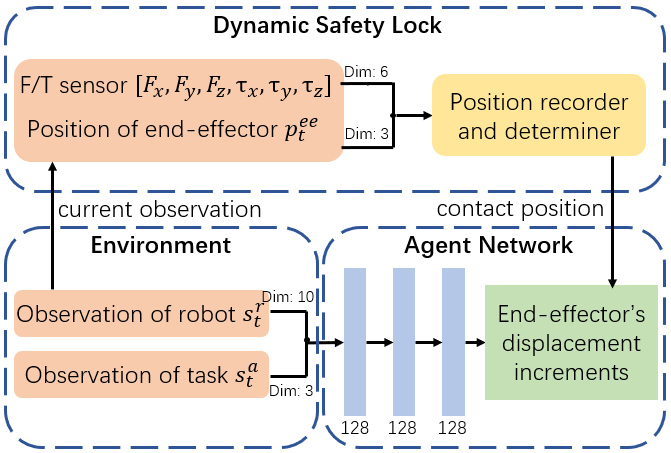}
    \caption{Network structure of robot assembly models. The entire network is divided into three parts, the environment, the agent network, and the dynamic safety lock (\ref{sec:sec:lock}).}
    \label{fig:net}
\end{figure}


\subsection{DSL for Robot}
\label{sec:sec:lock}

To ensure that the robot does not collide with the environment violently and thus has problems such as system crashes, as shown in Fig. \ref{fig:net}, we suggest a DSL method in each control loop. 
The vector consisting of these two sets of signals is used as the input to the DSL. 
When the EEF touches around the hole, the change in the value of the F/T sensor becomes larger and the recorded position $\mathbf{R}_{p}=[x^{ee}_{ti}, y^{ee}_{ti}, z^{ee}_{ti}](i=1,2,3,...)$ at this moment is used as the limited contact position $z^{c}_{t}$ for the next time of exploration. 
To speed up the exploration to find the initial first limit, we artificially add a tiny increment $\delta z^{ee}_t$ in the direction of the insertion of the output increment, which does not affect the overall trajectory of the RL model exploration.
Finally, to more concretely represent the function of the DSL, we list the pseudo-algorithm as shown in Algorithm \ref{alg:lock}. 

\begin{figure}
    \centering
    \includegraphics[width=0.35\textwidth]{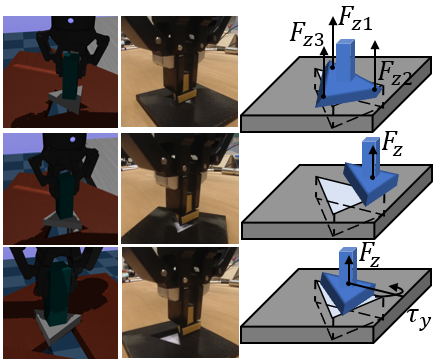}
    \caption{Schematic diagram of the detection points for setting the dynamic safety lock. Each line is a detection behavior, respectively, simulation, real-world and geometry schematic. Each contact point represents a small contact zone, and the contact forces, where $F_z = F_{z1}+F_{z2}+F_{z3}$, $\tau_y$ represents the torque in the y-direction, are represented by the F/T sensor.}
    \label{fig:insertpose}
\end{figure}

Since exploring the edge of the hole or not has a relatively large effect on the F/T sensor as shown in Fig. \ref{fig:insertpose}, we set threshold $\mathbf{\delta f}$, which is activated when the edge of the hole is explored on the first exploration, to be used to broaden the limit $z^c_t$. The specific calculation is as follows.
\begin{equation}
\begin{aligned}
\delta x^{ee}_t = \beta_{11}\cdot(F_{x(t)} - F_{x(t-1)} + \tau_{x(t)} - \tau_{x(t-1)})\\
\delta y^{ee}_t = \beta_{12}\cdot(F_{y(t)} - F_{y(t-1)} + \tau_{y(t)} - \tau_{y(t-1)})\\
\delta z^{ee}_t = \beta_{13}\cdot(F_{z(t)} - F_{z(t-1)} + \tau_{z(t)} - \tau_{z(t-1)}),
\end{aligned}
\label{eqn:f2z}
\end{equation}
where $F_{\cdot(t)}$, $F_{\cdot (t-1)}$, $\tau_{\cdot(t)}$ and $\tau_{\cdot (t-1)}$ denote the last two values of the $\mathbf{R}_{f}$ record. $\mathbf{\beta_1}= [\beta_{11}, \beta_{12}, \beta_{13}]$ represents the F/T variation gain vector. The tiny increments $\delta x^{ee}_t$, $\delta y^{ee}_t$ and $\delta z^{ee}_t$ obtained are used as leverage to raise the limit $z^c_t$.
\begin{equation}
\begin{aligned}
z^c_t = z^{ee}_t + \delta x^{ee}_t + \delta y^{ee}_t + \delta z^{ee}_t,
\end{aligned}
\label{eqn:zctRf}
\end{equation}
where $z^{ee}_t$ denotes the last value of the $\mathbf{R}_{p}$ record. On the other hand, if the threshold R is not activated, i.e., it is considered that the edge of the hole is not touched, but the strength of the last position change $\delta\mathbf{R}_p$ reflects the intensity of the downward exploration, so $\delta\mathbf{R}_p$ is used as a tiny increment to regulate the limit $z^c_t$.
\begin{equation}
\begin{aligned}
\delta\mathbf{R}_p = \Vert \mathbf{R}_{p(t)} - \mathbf{R}_{p(t-1)} \Vert,
\end{aligned}
\label{eqn:Rpchange}
\end{equation}
\begin{equation}
\begin{aligned}
z^{c}_t = z^{ee}_t + \mathbf{\beta_2}\cdot \delta\mathbf{R}_{p},
\end{aligned}
\label{eqn:zctRp}
\end{equation}
where $\mathbf{R}_{p(t)}$ and $\mathbf{R}_{p(t-1)}$ denote the last value it record. $\mathbf{\beta_2}$ represents the gain vector of $\delta\mathbf{R}_p$.

\begin{algorithm}[t]
	\caption{DSL.}
	\KwIn{\\
            \begin{itemize}
            \item[$\bullet$] F/T sensor: $[F_x, F_y, F_z, \tau_x, \tau_y, \tau_z]$
            \item[$\bullet$]Position of the EEF: $[x^{ee}_t, y^{ee}_t, z^{ee}_t]$
            \end{itemize}
            }
	\KwOut{Contact position $z^{c}_t$ from the recorder.} 
        \BlankLine
        // the observation can be provided for this algorithm.\\
	Initialize $p^h_t$, $p^{ee}_t$ randomly\\
        Define tiny increment $\delta z^{ee}_t$\\
        Define the six-dimensional force threshold $\mathbf{\delta f}$\\
        Normalized the F/T sensor data\\
	\While{\textnormal{not contact ($F_z < 0.5$)}}{
		Record F/T sensor's data $[F_x, F_y, F_z, \tau_x, \tau_y, \tau_z]$ and position $[x^{ee}_t, y^{ee}_t, z^{ee}_t]$ as $\mathbf{R}_{f}$, $\mathbf{R}_p$\\
        Add tiny increment $\delta z^{ee}_t$ into $z^{ee}_t$ in the direction of insertion to gradually explore downward. \\ 
	}
        
        \eIf{$\mathbf{R}_f(t) - \mathbf{R}_f(t-1) > \mathbf{\delta f}$ }{
        Utilize changes in F/T sensor to obtain the tiny increments $\delta x^{ee}_t, \delta y^{ee}_t, \delta z^{ee}_t$ (Equation (\ref{eqn:f2z}))\\
        Tiny increments act as gains on $z^c_t$ (Equation (\ref{eqn:zctRf}))\\
      }{
      Calculate the intensity of the position $\delta\mathbf{R}_p$ change at the last moment (Equation (\ref{eqn:Rpchange})) \\
      Tiny increments $\delta\mathbf{R}_p$ act as gains on $z^c_t$ (Equation (\ref{eqn:zctRp}))\\
      }
 \label{alg:lock}
\end{algorithm}

\section{Experiments and results}
\label{sec:experiment}
\vspace{-5pt}


\subsection{Simulation Experiments}
\label{sec:sec:sim}

\subsubsection{Experimental Setup Details}
\label{sec:sec:sec:sim-setup}
All training and testing in the simulation part are on the Intel(R) Core(TM) i7-1185G7 CPU. For the hyper-parameters mentioned throughout this paper, this work tries several times to obtain the best hyper-parameters we could get. 
For the reward function (Equation (\ref{eqn:rewardfunc})), $\alpha_1=\alpha_2=2.30$, $\alpha_3=1.23$, $\alpha_4=2$, $\alpha_5=0.5$, $\delta_1=1\mathrm{e}{-04}$, $\delta_2=0.01$ (Equation (\ref{eqn:zdist})). 
For the DSL, $\mathbf{\beta_1}=[1\mathrm{e}{-03}, 1\mathrm{e}{-03}, 5\mathrm{e}{-04}]$, $\mathbf{\beta_2}=[1\mathrm{e}{-07}, 1\mathrm{e}{-07}, 1\mathrm{e}{-03}]$, $\mathbf{\delta f}=[0.15, 0.15, 0.45, 0.1, 0.1, 0.2]$. 
For the observation, the Gaussian noise 
is added to the observations of the holes. The observation of the robot $s^{r}_{t}$ has the noise of the simulator, so no additional noise is added to it.
The total number of training steps is $2\mathrm{e}{06}$, and the maximum number of steps per episode is $110$. 
For the task, we randomize the configuration of the hole position and orientation at the beginning of each episode to enhance the robustness and generalization of the model. All models are set with checkpoints and estimated models are generated every $1\mathrm{e}{04}$ steps. The estimated models are set up to test the success rate of the trained models, which is judged by the bottom of the object being more than $2.5 mm$ below the surface of the hole.



\subsubsection{Vision and F/T Model}
\label{sec:sec:pihtask}
We set up three sets of experiments with different data inputs, which constitute three models with DSL, the vision-based model (VM), the vision-F/T-based model (VFTM), and the F/T-based model (FTM).

\begin{figure}
    \centering
    \includegraphics[width=0.5\textwidth]{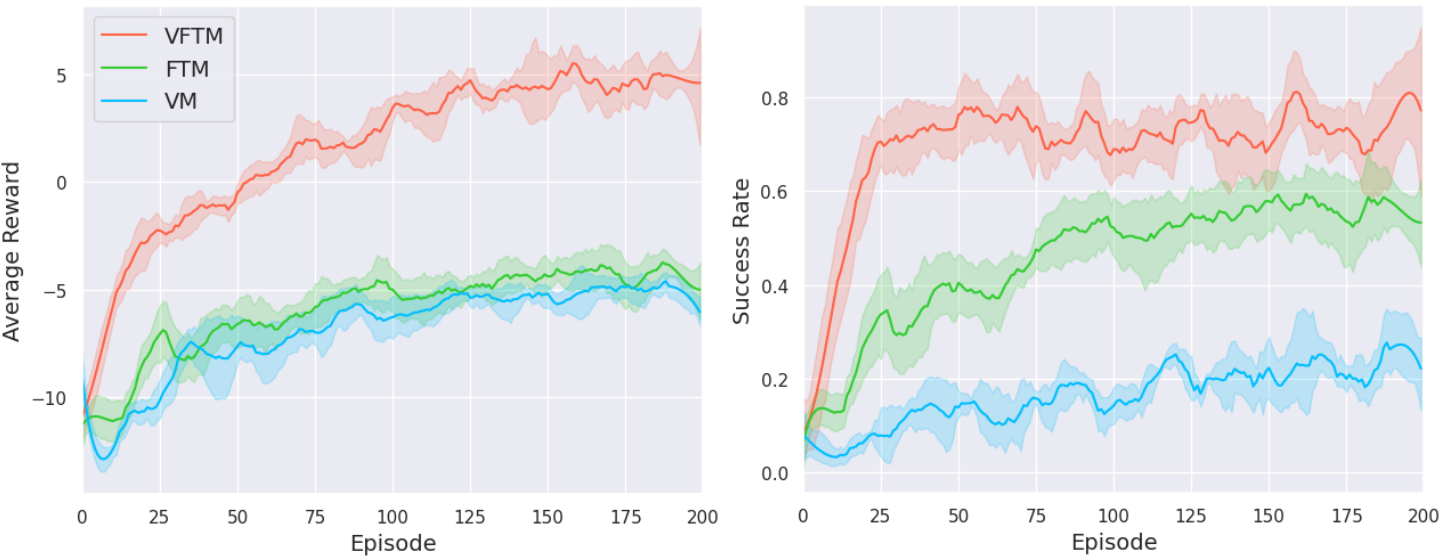}
    \caption{Training performance of three assembly models ($\mathrm{VFTM}$, $\mathrm{FTM}$ and $\mathrm{VM}$) based on vision and F/T sensor with the fixed gap. }
    \label{fig:pih4mm3models}
\end{figure}

We set a single fixed assembly gap with a size of $4.0mm$ and a fixed size of the grasped object. As shown in Fig. \ref{fig:pih4mm3models}, firstly, for the $\mathrm{VM}$, we set the $\beta_2=0$ to disable the DSL. Then, we can see that $\mathrm{VFTM}$ has the highest reward with the fixed gap and has a high success rate. The success rate performance of $\mathrm{FTM}$ is better than $\mathrm{VM}$ one. A reason is that we do not use a camera-based shape recognition algorithm that requires a large amount of data to identify the shape and orientation of the hole, but rather randomize the position and orientation of the hole, while $\mathrm{FTM}$ is able to learn the skill of alignment through the encoding of F/T sensor data. However, as this model has no visual support, $\mathrm{FTM}$ does not know where the hole is. Thus, $\mathrm{FTM}$ can only get the position of the hole by constantly exploring, which leads to sometimes when the entire episode is over did not explore the position of the hole so that the success rate is not remarkable.

Finally, we take the best result of each model as the best model for testing and get the results as shown in Table \ref{tab:modelidentify}, where each model is tested $500$ times independently to obtain the reward and success rate. In this subpart, we focus on the $4mm$ gap and we can see that the success rates of $\mathrm{VFTM}$, $\mathrm{FTM}$ and $\mathrm{VM}$ are $88.57\%$, $56.42\%$ and $25.59\%$, respectively, which are consistent with the results analyzed from Fig. \ref{fig:pih4mm3models} in the above. For the reward, $\mathrm{FTM}$ has the largest variance, which is reasonable because of its need for continuous exploration. 

It can be proven that the model with rich data is more effective, the F/T sensor can replace the shape recognition function of the visual sensor to accomplish the task.

\begin{table}[htbp]
\vspace{3mm}
\begin{center}
\caption{\textcolor{black}{Means (ME) and variances (VAR) of model testing in simulation. (r: reward, sr: success rate)}}
\begin{tabular}{|cc|cc|cc|}
\hline
\multicolumn{2}{|c|}{\multirow{2}{*}{}}                  & \multicolumn{2}{c|}{4mm}              & \multicolumn{2}{c|}{1mm}              \\ \cline{3-6} 
\multicolumn{2}{|c|}{}                                   & \multicolumn{1}{c|}{ME}      & VAR    & \multicolumn{1}{c|}{ME}      & VAR    \\ \hline
\multicolumn{1}{|c|}{\multirow{2}{*}{VFTM-DSL}}     & r  & \multicolumn{1}{c|}{4.7399}  & 1.0407 & \multicolumn{1}{c|}{-0.3211} & 0.4442 \\ \cline{2-6} 
\multicolumn{1}{|c|}{}                              & sr & \multicolumn{1}{c|}{0.8857}  & 0.0043 & \multicolumn{1}{c|}{0.4195}  & 0.0024 \\ \hline
\multicolumn{1}{|c|}{\multirow{2}{*}{VFTM-sliding}} & r  & \multicolumn{1}{c|}{-4.2873} & 0.0309 & \multicolumn{1}{c|}{-4.2904} & 0.0048 \\ \cline{2-6} 
\multicolumn{1}{|c|}{}                              & sr & \multicolumn{1}{c|}{0.2503}  & 0.0027 & \multicolumn{1}{c|}{0.0064}  & 0.5352 \\ \hline
\multicolumn{1}{|c|}{\multirow{2}{*}{FTM-DSL}}      & r  & \multicolumn{1}{c|}{-4.5648} & 0.6895 & \multicolumn{1}{c|}{-}       & -      \\ \cline{2-6} 
\multicolumn{1}{|c|}{}                              & sr & \multicolumn{1}{c|}{-0.5642} & 0.0033 & \multicolumn{1}{c|}{-}       & -      \\ \hline
\multicolumn{1}{|c|}{\multirow{2}{*}{VM}}           & r  & \multicolumn{1}{c|}{-5.3125} & 0.2304 & \multicolumn{1}{c|}{-}       & -      \\ \cline{2-6} 
\multicolumn{1}{|c|}{}                              & sr & \multicolumn{1}{c|}{0.2559}  & 0.0013 & \multicolumn{1}{c|}{-}       & -      \\ \hline
\end{tabular}
\label{tab:modelidentify}
\end{center}
\vspace{-5mm}
\end{table}

\subsubsection{DSL Experiment}
\label{sec:sec:pihdsl}
This part, at last, establishes ablation experiments about the proposed DSL method. As mentioned in \ref{sec:sec:lock}, the DSL method differs from the traditional sliding one by having a repetitive pressing action. Therefore, this experiment only distinguishes the effects of different action styles (DSL vs. Sliding). From the above experimental results, it is obtained that $\mathrm{VFTM}$ is the most effective, so this experiment is conducted with $\mathrm{VFTM}$ at different sizes of gaps. 

\begin{figure}
    \centering
    \includegraphics[width=0.5\textwidth]{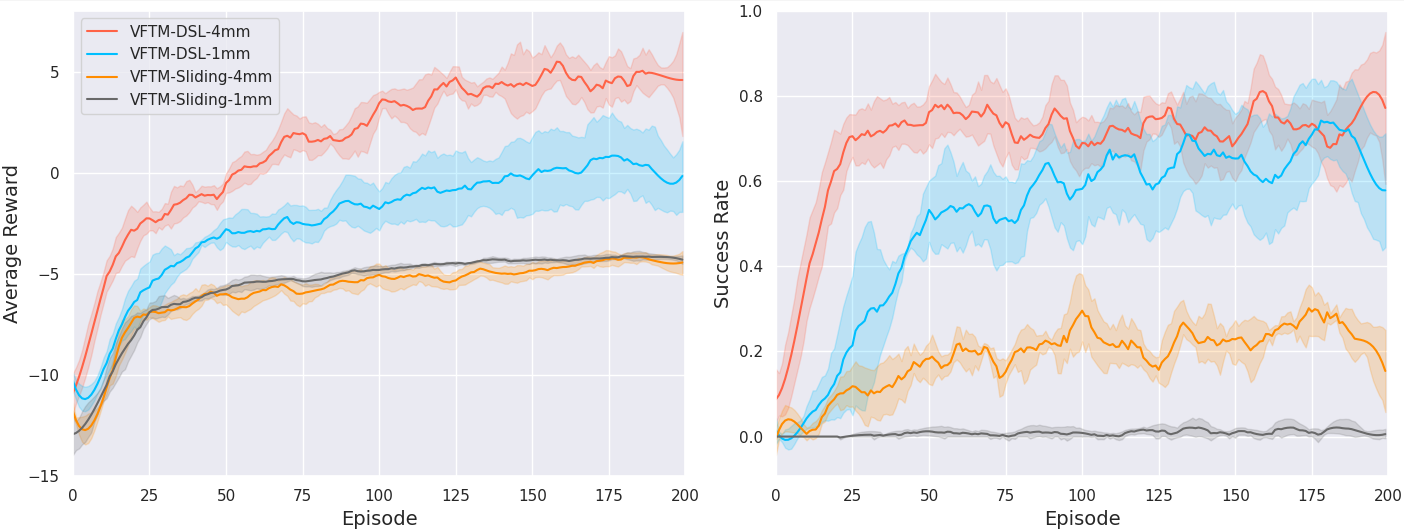}
    \caption{Performance of training method (DSL vs. Sliding).}
    \label{fig:dsl}
\end{figure}

\begin{table*}[]
\vspace{3mm}
\begin{center}
\caption{\textcolor{black}{Insertion experiment results.}}
\begin{tabular}{|c|cc|c|cc|c|cc|c|}
\hline
             & \multicolumn{2}{c|}{tr}                & rtr     & \multicolumn{2}{c|}{trm}               & cir     & \multicolumn{2}{c|}{b-trm}             & b-rtr   \\ \hline
proportion   & \multicolumn{1}{c|}{26.3\%}  & 7.8\%   & 24.7\%  & \multicolumn{1}{c|}{30.65\%} & 9.33\%  & 13.78\% & \multicolumn{1}{c|}{20.50\%} & 5.8\%   & 5.5\%   \\ \hline
success rate & \multicolumn{1}{c|}{79.63\%} & 18.64\% & 68.52\% & \multicolumn{1}{c|}{83.67\%} & 15.43\% & 20.00\% & \multicolumn{1}{c|}{41.37\%} & 13.95\% & 22.81\% \\ \hline
\end{tabular}
\label{tab:realtab}
\end{center}
\vspace{-5mm}
\end{table*}

As shown in \textcolor{black}{Fig. \ref{fig:dsl}}, 
qualitatively, for the reward, the obtained score by the DSL method is higher than the sliding method one for each size of the gap. For the success rate, The performance of both sliding methods is low, while the performance of the DSL method is remarkable. 
Quantitatively, as shown in Table \ref{tab:modelidentify}, the data corresponding to Fig. \ref{fig:dsl} in this subpart are $\mathrm{VFTM}$ with DSL method and $\mathrm{VFTM}$ with sliding method.
The success rate of the sliding method with the $4mm$ gap is $25.03\%$ but the DSL method has $88.57\%$, which indicates that the sliding method does not match the hole well at the smaller gap. For the fluctuation of the data, i.e., the variance, the fluctuation of the sliding method is smaller compared to the DSL method, which indicates that the method repeatedly explores around the hole location after exploring it and does not find the insertion direction.


\subsection{Real-World Robotics Experiments}
\label{sec:sec:real}
\begin{figure}
    \centering
    \includegraphics[width=0.25\textwidth]{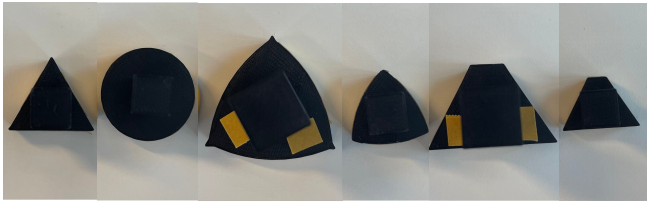}
    \caption{Objects of different shapes and sizes, from left to right are triangle (tr), circle (cir), big size Reuleaux triangle (b-rtr), Reuleaux triangle (rtr) and triangle with missing corner (trm) and the big size one (b-trm).}
    \label{fig:shapes}
\end{figure}

The experiments are completed on the device UR3e+Robotiq2F. The hole position with the marker is captured by a ZED2i camera.
Since the effectiveness of the DSL method has been demonstrated in the simulation, we no longer compare the DSL method and the sliding method in the real world. According to the characteristics of the F/T sensor, we designed an exploratory experiment to generalize the model based on the triangle training in Fig. \ref{fig:insertpose} to other shapes or other sizes as shown in Fig.\ref{fig:shapes}. 

\begin{figure}
    \centering
    \includegraphics[width=0.4\textwidth]{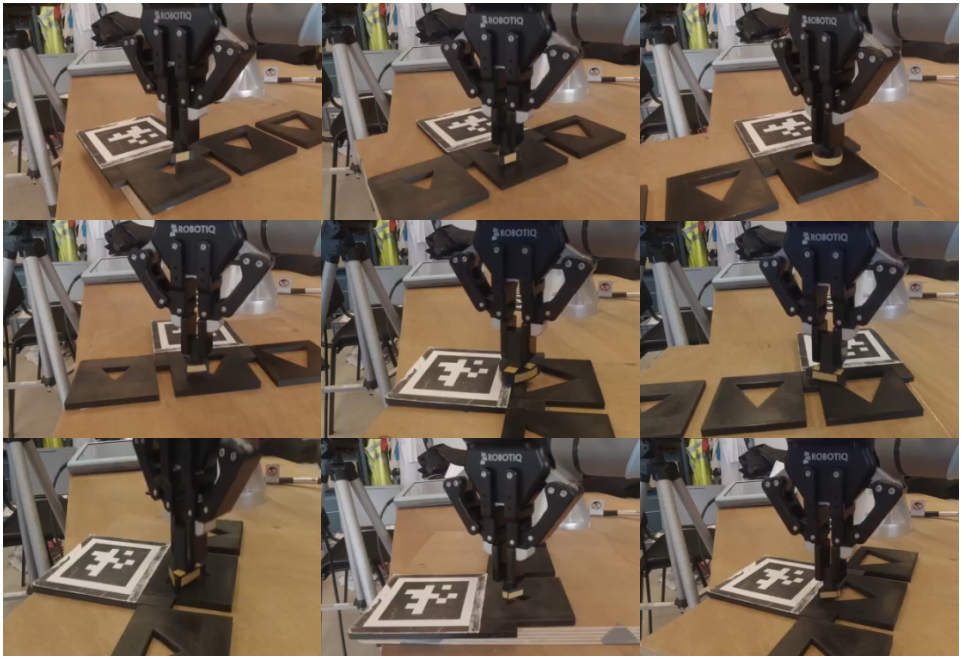}
    \caption{Insertion experiment, from top to bottom, from left to right are $tr$ with $7.8\%$ and $26.3\%$ proportion, $cir$ with $13.78\%$ proportion, $rtr$ with $24.7\%$ proportion, $b$-$rtr$ with  $5.5\%$ proportion, $b$-$trm$ with $20.50\%$ and $5.8\%$ proportion, $trm$ with $9.33\%$ and $30.65\%$ proportion.}
    \label{fig:realexpdiff}
\end{figure}

Based on Fig.\ref{fig:shapes}, we also changed the size of the hole and used the area proportion of the object to the hole instead of the gap size ($4mm$) to describe the gap. The smaller the proportion, the smaller the gap. As shown in Fig. \ref{fig:realexpdiff}, we conduct experiments with the objects shown in Fig. \ref{fig:shapes} and holes of different sizes. Note that all experiments are performed with the same model (VFTM-DSL), the purpose is to test the generalization effect of the model by using the perceptual ability of the F/T sensor. 

We tested different situations about 50 times, each time artificially changing the orientation and position of the hole randomly, and got the results shown in Table \ref{tab:realtab}. It can be seen that the success rate of $tr$ (original model) is $79.63\%$, which is similar to the test results in simulation, and it has a success rate of $28.64\%$ when the gap is $1mm$ ($7.8\%$ proportion). Objects of similar shape ($rtr$, $trm$) can have a higher success rate under similar size conditions ($68.52\%$, $83.67\%$ and $25.43\%$). In particular, the proportion of $trm$ is larger than the $tr$ one, and its success rate is higher than the original model one. For the object ($cir$) with a large difference in shape, the success rate is reduced to $20\%$. This is because the force sensor failed to find a suitable pose to insert during the pressing process due to the mismatch of the object shape. For objects of different sizes ($b$-$trm$, $b$-$rtr$), the success rates ($31.37\%$, $13.95\%$ and $22.81\%$) are reduced compared to the original model. This is because when the object is larger, the change of the F/T sensor's value is reduced during the pressing process, thereby reducing the probability of exploring the hole. However, there is still a success rate ($31.37\%$) when the gap proportion is large ($20.50\%$).

In short, we use the F/T sensor to replace some of the functions of the camera and use the changing characteristics of the sensor's feedback during the exploration process, so that it can be applied to objects of similar shape and size.

\bibliographystyle{ieeetr}
\bibliography{ICRA_Demo}

\end{document}